\documentclass[10pt,twocolumn,letterpaper]{article}

\usepackage[pagenumbers]{style} 

\newcommand{\TODO}[1]{}

\usepackage[T1]{fontenc}
\usepackage{booktabs}
\usepackage{multirow}
\usepackage{enumitem}
\usepackage{amsmath}
\usepackage{amssymb}
\usepackage{needspace}
\usepackage{tikz}
\usetikzlibrary{arrows.meta, positioning, fit, backgrounds, calc}

\newcommand{\ci}[1]{{\scriptsize$\pm$#1}}

\usepackage{microtype}

\definecolor{linkblue}{rgb}{0.21,0.49,0.74}
\usepackage[breaklinks,hidelinks,colorlinks]{hyperref}
\hypersetup{
  linkcolor=linkblue,
  citecolor=linkblue,
  urlcolor=linkblue
}

\def\paperID{} 
\def\confName{}
\def\confYear{2026}

\title{Geometry-Aware Metric Learning for Cross-Lingual\\Few-Shot Sign Language Recognition on Static Hand Keypoints}

\author{Chayanin Chamachot\\
Department of Computer Engineering\\
Faculty of Engineering, Chulalongkorn University
\and
Kanokphan Lertniphonphan\\
Department of Computer Engineering\\
Faculty of Engineering, Chulalongkorn University
}

\begin{document}
\maketitle
\begin{abstract}
Sign-language recognition (SLR) systems typically require large labelled corpora for each language, yet the majority of the world's 300{+} sign languages lack sufficient annotated data.
Cross-lingual few-shot transfer---pretraining on a data-rich source language and adapting with only a handful of target-language examples---offers a scalable alternative, but conventional coordinate-based keypoint representations are susceptible to domain shift arising from differences in camera viewpoint, hand scale, and recording conditions.
This shift is particularly detrimental in the few-shot regime, where class prototypes estimated from only $K$ examples are highly sensitive to extrinsic variance.
We propose a geometry-aware metric-learning framework centred on a compact 20-dimensional inter-joint angle descriptor derived from MediaPipe static hand keypoints.
These angles are provably invariant to $\mathrm{SO}(3)$ rotation, translation, and isotropic scaling, thereby eliminating the dominant sources of cross-dataset shift and yielding tighter, more stable class prototypes.
Evaluated on four fingerspelling alphabets spanning typologically diverse sign languages---ASL, LIBRAS, Arabic~SL, and Thai---the proposed angle features improve over normalised-coordinate baselines by up to 25 percentage points within-domain and enable frozen cross-lingual transfer that frequently exceeds within-domain accuracy, all using a lightweight MLP encoder (${\sim}$105\,k parameters).
These findings demonstrate that formally invariant hand-geometry descriptors provide a portable and effective foundation for cross-lingual few-shot SLR in low-resource settings.\footnote{Code and data splits are available at \url{https://github.com/fjkrch/sign_metric_learning}.}
\end{abstract}    
\section{Introduction}
\label{sec:intro}

Sign language is the primary mode of communication for over 70 million Deaf individuals worldwide~\cite{who2024deafness}, yet recognition technology currently serves only a small fraction of the 300{+} documented sign languages~\cite{ethnologue2024}.
Building an SLR system for an under-resourced language typically requires thousands of labelled examples per class---a requirement that is prohibitively expensive for most communities.
This data bottleneck motivates \emph{cross-lingual few-shot transfer}: pretraining a model on a data-rich source language and adapting it to a new target language using only a handful of labelled examples.

\paragraph{Challenge: domain shift in keypoint representations.}
State-of-the-art continuous SLR systems fuse RGB, optical flow, and body/hand keypoints~\cite{camgoz2020sign,li2020word}, but they depend on large video corpora and learn language-specific patterns.
Existing isolated SLR benchmarks evaluate single-language, closed-set classification~\cite{mavi2020asl,podder2022libras,alani2021arabic}, leaving it unclear whether learned representations generalise across languages.
Although cross-lingual transfer is well studied in NLP~\cite{conneau2020xlmr} and speech~\cite{li2022massively}, it remains largely unexplored for sign languages; Tavella~\etal~\cite{tavella2022fewshot} apply few-shot learning to signs but only within a single language.
A key obstacle is that normalised $(x,y,z)$ keypoint coordinates remain sensitive to camera viewpoint and hand scale, inducing domain shift across datasets collected under different conditions.
In the few-shot regime this problem is amplified: each Prototypical Network centroid is estimated as a sample mean over only $K$ examples, so extrinsic variance in the input directly inflates prototype estimates and destabilises classification.

\paragraph{Our approach.}
We propose a \emph{transferable geometric inductive bias}: a set of 20 inter-joint angles computed from MediaPipe~\cite{zhang2020mediapipe} static hand keypoints, provably invariant to rotation, translation, and isotropic scaling (\cref{sec:theory}).
\Cref{fig:pipeline} illustrates the full pipeline: a hand image is processed by MediaPipe into 21 keypoints, converted to one of three representations (\texttt{raw}, \texttt{angle}, or \texttt{raw\_angle}), and encoded into a 128-dimensional embedding for Prototypical Network classification.
Using a lightweight MLP encoder (${\sim}$105\,k parameters), we evaluate on four typologically diverse fingerspelling alphabets---\textbf{ASL} (29 classes), \textbf{LIBRAS} (21), \textbf{Arabic SL} (31), and \textbf{Thai} (42)---under a deterministic 5-way $K$-shot protocol.
Because the angle embedding is intrinsically invariant to rigid transforms, it requires no spatial normalisation across camera setups, is more privacy-preserving than RGB pipelines (only keypoints and angles are stored), and can serve as a plug-in pose branch in multimodal architectures.

\needspace{8\baselineskip}
\paragraph{Contributions.}
\begin{enumerate}[leftmargin=*,nosep]
    \item \textbf{Cross-lingual few-shot benchmark.}  We establish a deterministic 5-way $K$-shot evaluation protocol spanning four typologically diverse fingerspelling alphabets.  Angle features frequently match or exceed within-domain baselines under cross-lingual transfer (\cref{tab:within,tab:cross,tab:source_compare}).
    \item \textbf{Geometry-invariant representation.}  We derive a 20-dimensional inter-joint angle feature that is provably invariant to rotation, translation, and isotropic scaling (\cref{eq:invariance}).  Removing normalisation degrades \texttt{raw} coordinates by ${\sim}$5\,pp, while angle features remain unchanged ($|\Delta|\leq 0.3$\,pp; \cref{tab:ablation,tab:pipeline}).
    \item \textbf{Systematic baselines.}  We benchmark few-shot accuracy against input-space, episode-linear, full-data, and multi-seed baselines (\cref{tab:baselines_ext}), quantifying the cost of learning from only $K$ examples.
\end{enumerate}

\paragraph{Scope.}
This work evaluates isolated, static fingerspelling---not continuous SLR---using keypoints only, and is complementary to video-based approaches.

\section{Related Work}
\label{sec:related}

We organise prior work into four areas: skeleton-based representations, single-language SLR, few-shot learning for sign recognition, and cross-lingual or zero-shot transfer.

\paragraph{Skeleton and keypoint-based representations.}
MediaPipe Hands~\cite{zhang2020mediapipe} provides real-time 21-keypoint hand tracking, enabling lightweight SLR pipelines that typically flatten $(x,y,z)$ coordinates into a 63-dimensional vector~\cite{de2023sign,shin2023korean}.
Jiang~\etal~\cite{jiang2021skeleton} demonstrate that skeleton features generalise better than appearance-based features across recording conditions; however, they do not evaluate cross-lingual transfer.
Angle-based descriptors have a long history in action recognition~\cite{shi2019twostream,chen2021ctrgcn} and glove-based gesture systems~\cite{cheok2019review}, yet existing MediaPipe SLR pipelines have not exploited them.
Our 20-dimensional angle representation (\cref{sec:theory}) achieves exact $\mathrm{SO}(3)$ invariance under similarity transforms (\cref{eq:invariance}), enabling stable cross-dataset transfer without ad-hoc normalisation.

\paragraph{Single-language sign language recognition.}
CNN-based methods achieve over 99\% accuracy on the ASL Alphabet benchmark~\cite{mavi2020asl}; hand-crafted skeleton features reach 97\% on LIBRAS~\cite{podder2022libras} and Arabic SL~\cite{alani2021arabic}; graph convolutional networks further advance continuous SLR~\cite{li2020word}.
All of these approaches are \emph{single-language}: they require large per-class training sets and do not address cross-lingual generalisation.

\paragraph{Few-shot learning for sign recognition.}
Prototypical Networks~\cite{snell2017prototypical} learn a metric embedding space in which classification reduces to nearest-centroid matching.
Chen~\etal~\cite{chen2019closer} demonstrate that a well-tuned ProtoNet often matches more complex meta-learning alternatives such as MAML~\cite{finn2017maml}, Matching Networks~\cite{vinyals2016matching}, and Relation Networks~\cite{sung2018relation}.
Tavella~\etal~\cite{tavella2022fewshot} apply few-shot learning to sign language but evaluate only within a single language, leaving the cross-lingual setting unexplored.

\paragraph{Cross-lingual and zero-shot SLR.}
Boh\'{a}\v{c}ek~\cite{bohacek2023pose} explores pose-based few-shot SLR on ASLLVD-Skeleton but is limited to a single language.
Bilge~\etal~\cite{bilge2024cross} investigate zero- and few-shot SLR across ASL, DGS, and TID but rely on fully supervised target-language training.
Our work differs from both in three respects: (i)~we adopt a deterministic $N$-way $K$-shot episodic protocol, (ii)~we introduce formally invariant geometric features that remove extrinsic variance at the representation level, and (iii)~we evaluate cross-lingual transfer across four typologically diverse languages.

\section{Method}
\label{sec:method}

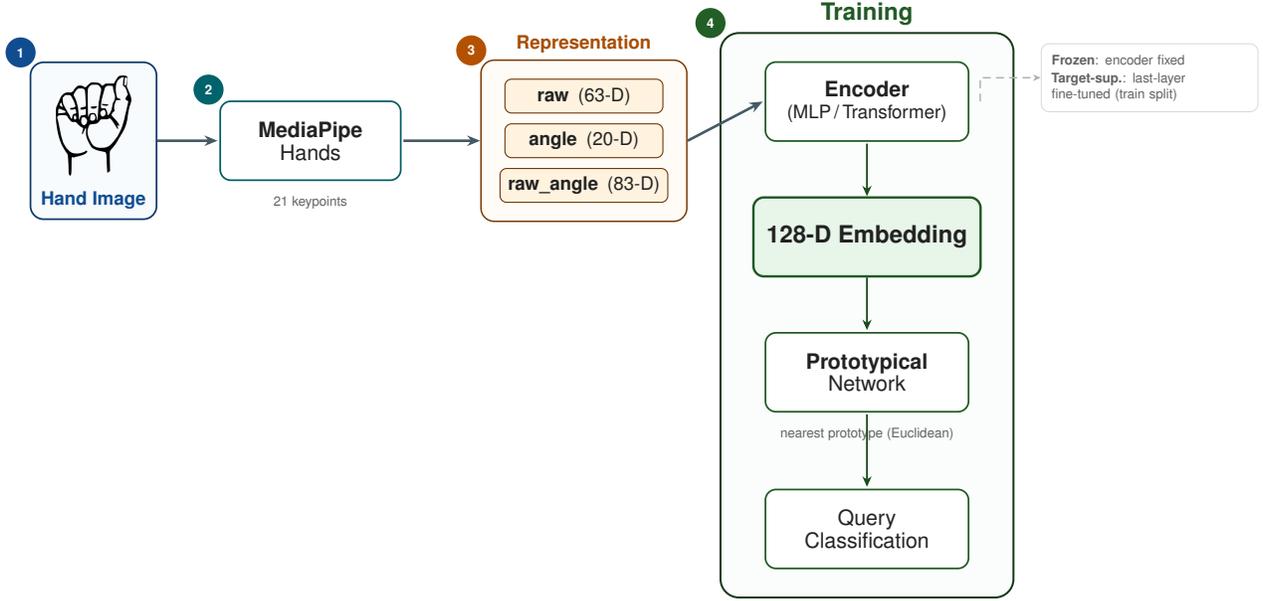
\begin{figure*}[t]
  \centering
  \definecolor{accentblue}{HTML}{1565C0}%
  \definecolor{accentteal}{HTML}{00838F}%
  \definecolor{accentamber}{HTML}{EF6C00}%
  \definecolor{accentgreen}{HTML}{2E7D32}%
  \definecolor{fillblue}{HTML}{E3F2FD}%
  \definecolor{fillteal}{HTML}{E0F7FA}%
  \definecolor{fillamber}{HTML}{FFF3E0}%
  \definecolor{fillgreen}{HTML}{E8F5E9}%
  \definecolor{darktext}{HTML}{212121}%
  \definecolor{midtext}{HTML}{616161}%
  \definecolor{arrowcolor}{HTML}{455A64}%
  \begin{tikzpicture}[
    >=Stealth,
    every node/.style={text=darktext},
    procblock/.style={
        draw=#1!70!black, line width=0.65pt,
        rounded corners=4pt, minimum height=1.05cm,
        minimum width=2.4cm, align=center,
        font=\sffamily\footnotesize, fill=white,
        inner sep=5pt, outer sep=1pt},
    procblock/.default=accentblue,
    repritem/.style={
        draw=accentamber!45!black, line width=0.4pt,
        rounded corners=2.5pt, minimum width=2.1cm, minimum height=0.45cm,
        align=center, font=\sffamily\scriptsize,
        fill=fillamber, inner sep=3pt},
    embblock/.style={
        draw=accentgreen!70!black, line width=0.9pt,
        rounded corners=4pt, minimum height=1.05cm,
        minimum width=2.7cm, align=center,
        font=\sffamily\small\bfseries, fill=fillgreen,
        inner sep=5pt},
    arr/.style={-{Stealth[length=5pt, width=4pt]},
        line width=1.0pt, draw=arrowcolor},
    inarr/.style={-{Stealth[length=4pt, width=3.5pt]},
        line width=0.7pt, draw=accentgreen!60!black},
    dasharr/.style={-{Stealth[length=3pt, width=2.5pt]},
        line width=0.55pt, draw=midtext!50, densely dashed},
    stagelabel/.style={font=\sffamily\scriptsize\bfseries,
        text=#1!70!black},
    stagelabel/.default=accentblue,
    sublabel/.style={font=\sffamily\tiny, text=midtext},
    badge/.style={circle, minimum size=0.4cm, inner sep=0pt,
        font=\sffamily\tiny\bfseries, text=white,
        fill=#1!75!black},
    badge/.default=accentblue,
    annbox/.style={font=\sffamily\tiny, text=midtext, align=left,
        draw=midtext!25, rounded corners=3pt, fill=white,
        inner sep=4pt, line width=0.35pt, text width=2.6cm},
  ]
    \node[draw=accentblue!55!black, line width=0.65pt, rounded corners=5pt,
          fill=fillblue!30, inner sep=4pt, align=center] (img) {%
      \includegraphics[height=1.4cm]{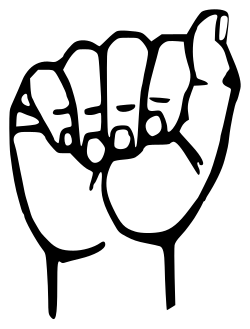}\\[-2pt]
      {\sffamily\scriptsize\color{accentblue!75!black}\bfseries Hand Image}};
    \node[badge=accentblue]
      at ([shift={(-0.12,0.12)}]img.north west) {1};

    \node[procblock=accentteal, right=0.8cm of img] (mp)
      {{\bfseries MediaPipe}\\[-1pt]Hands};
    \node[sublabel, below=0.06cm of mp] (mplbl) {21 keypoints};
    \node[badge=accentteal]
      at ([shift={(-0.12,0.12)}]mp.north west) {2};

    \coordinate (rcenter) at ($(mp.east)+(2.4cm,0)$);
    \node[repritem] at (rcenter)               (r2) {\textbf{angle}\enspace(20-D)};
    \node[repritem, above=0.13cm of r2]        (r1) {\textbf{raw}\enspace(63-D)};
    \node[repritem, below=0.13cm of r2]        (r3) {\textbf{raw\_angle}\enspace(83-D)};
    \begin{scope}[on background layer]
      \node[draw=accentamber!50!black, line width=0.55pt, rounded corners=5pt,
            fill=fillamber!25, inner sep=7pt, fit=(r1)(r2)(r3),
            label={[stagelabel=accentamber, anchor=south, yshift=-1pt]%
                    above:Representation}] (repr) {};
    \end{scope}
    \node[badge=accentamber]
      at ([shift={(-0.12,0.12)}]repr.north west) {3};

    \node[procblock=accentgreen, minimum width=2.7cm,
          anchor=north west,
          at={($(repr.north east)+(1.0cm,0)$)}] (enc)
      {{\bfseries Encoder}\\[-1pt]{\scriptsize(MLP\,/\,Transformer)}};
    \node[embblock, below=0.7cm of enc] (emb) {128-D Embedding};
    \node[procblock=accentgreen, minimum width=2.7cm, below=0.7cm of emb] (proto)
      {{\bfseries Prototypical}\\[-1pt]Network};
    \node[sublabel, below=0.05cm of proto] (protolbl)
      {nearest prototype (Euclidean)};
    \node[procblock=accentgreen, minimum width=2.7cm, below=0.5cm of protolbl] (cls)
      {Query\\[-1pt]Classification};
    \begin{scope}[on background layer]
      \node[draw=accentgreen!45!black, line width=0.75pt, rounded corners=7pt,
            fill=fillgreen!20, inner xsep=12pt, inner ysep=10pt,
            fit=(enc)(emb)(proto)(protolbl)(cls),
            label={[stagelabel=accentgreen, font=\sffamily\small\bfseries,
                    anchor=south, yshift=-1pt]above:Training}] (train) {};
    \end{scope}
    \node[badge=accentgreen]
      at ([shift={(-0.12,0.12)}]train.north west) {4};

    \draw[arr]   (img.east)  -- (mp.west);
    \draw[arr]   (mp.east)   -- (repr.west);
    \draw[arr]   (repr.east) -- (enc.west);
    \draw[inarr] (enc.south) -- (emb.north);
    \draw[inarr] (emb.south) -- (proto.north);
    \draw[inarr] (proto.south) -- ++(0,-0.22cm) -- (cls.north);

    \node[annbox, anchor=north west]
      at ([xshift=0.35cm, yshift=-0.15cm]train.north east) (anntext) {%
        \textbf{Frozen}: encoder fixed\\[1pt]
        \textbf{Target-sup.}: last-layer\\
        fine-tuned (train split)};
    \draw[dasharr] ([xshift=0.12cm]enc.east) |- (anntext.west);
  \end{tikzpicture}
  \caption{Pipeline overview.  A hand image is processed by MediaPipe Hands into 21 keypoints (63-D when flattened), converted to one of three representations---\texttt{raw}\,(63-D), \texttt{angle}\,(20-D), or \texttt{raw\_angle}\,(83-D)---and encoded into a 128-D embedding for Prototypical Network classification.  In the cross-lingual setting, the encoder is either \emph{frozen} or undergoes \emph{target-supervised adaptation} (last-layer fine-tuning on the target train split).}
  \label{fig:pipeline}
\end{figure*}

Our framework comprises four stages: (1)~extraction of 21 hand keypoints via MediaPipe, (2)~conversion to an $\mathrm{SO}(3)$-invariant angle descriptor (\cref{sec:theory}), (3)~encoding into a 128-dimensional embedding (\cref{sec:encoders}), and (4)~classification by nearest-prototype matching (\cref{sec:protocol}).  \Cref{fig:pipeline} illustrates the complete pipeline.

\subsection{Keypoints and Raw-Coordinate Preprocessing}
\label{sec:preprocess}

\begin{figure}[t]
  \centering
  \includegraphics[width=0.85\linewidth]{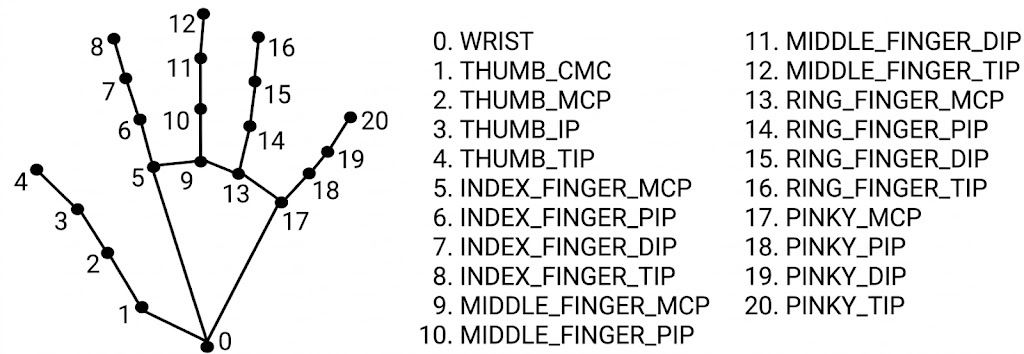}
  \caption{MediaPipe Hands keypoint topology ($i \in \{0,\ldots,20\}$).  Keypoint~0 is the wrist (root); each finger forms a four-joint kinematic chain following the official ordering~\cite{zhang2020mediapipe}: Thumb (1--4), Index (5--8), Middle (9--12), Ring (13--16), Pinky (17--20).}
  \label{fig:hand_topology}
\end{figure}

Given a hand image, we extract 21 three-dimensional landmarks $\{\mathbf{p}_i\}_{i=0}^{20}$ using MediaPipe Hands~\cite{zhang2020mediapipe}, where each landmark $\mathbf{p}_i = (x_i, y_i, z_i) \in \mathbb{R}^3$ represents the spatial position of a hand joint.
Keypoint $\mathbf{p}_0$ corresponds to the wrist and serves as the root of the hand skeleton.  The remaining 20 keypoints are arranged in five four-joint kinematic chains---Thumb (indices 1--4), Index (5--8), Middle (9--12), Ring (13--16), and Pinky (17--20)---as depicted in \cref{fig:hand_topology}.

The \textbf{raw} coordinate representation is obtained through two normalisation steps:
\begin{enumerate}[nosep,leftmargin=*]
    \item \textbf{Translation invariance (wrist-centring).}  The wrist position is subtracted from all keypoints to remove global translation:
    \begin{equation}
        \hat{\mathbf{p}}_i = \mathbf{p}_i - \mathbf{p}_0, \quad i \in \{0,\ldots,20\}.
        \label{eq:wrist_centre}
    \end{equation}
    \item \textbf{Scale invariance.}  All centred keypoints are divided by the maximum pairwise distance, removing dependence on hand scale and camera distance:
    \begin{equation}
        \tilde{\mathbf{p}}_i = \frac{\hat{\mathbf{p}}_i}{\displaystyle\max_{j,k \in \{0,\ldots,20\}}\|\hat{\mathbf{p}}_j - \hat{\mathbf{p}}_k\|}, \quad i \in \{0,\ldots,20\}.
        \label{eq:scale_norm}
    \end{equation}
\end{enumerate}
The resulting \textbf{raw} representation is obtained by row-major flattening of the normalised keypoints:
\begin{equation}
    \mathbf{x}_{\mathrm{raw}} = \mathrm{vec}\!\bigl(\tilde{\mathbf{p}}_0, \tilde{\mathbf{p}}_1, \ldots, \tilde{\mathbf{p}}_{20}\bigr) \in \mathbb{R}^{63},
    \label{eq:xraw}
\end{equation}
where $\mathrm{vec}(\cdot)$ denotes concatenation of all coordinate components and $63 = 21 \times 3$.
Importantly, the angle representation introduced next is computed from the \emph{original} (unnormalised) keypoints $\{\mathbf{p}_i\}$ and does not depend on the normalisation steps above.

\subsection{Geometry-Aware Angle Representation}
\label{sec:theory}

The 21-keypoint hand skeleton forms a kinematic tree rooted at the wrist ($\mathbf{p}_0$; see \cref{fig:hand_topology}).
Each non-wrist keypoint has a unique parent in this tree, producing 20 anatomical triplets $(a_k, j_k, c_k)$ for $k \in \{1, \ldots, 20\}$, where $a_k$, $j_k$, and $c_k$ denote the \emph{parent}, \emph{joint} (pivot), and \emph{child} keypoint indices, respectively.
Each triplet connects consecutive links along a finger chain.
As a concrete example, $(a_k, j_k, c_k){=}(5,6,7)$ measures flexion at the proximal interphalangeal (PIP) joint of the index finger, with keypoint~5 (MCP) as parent, 6 (PIP) as pivot, and 7 (DIP) as child.

For each triplet, we define displacement vectors from the pivot to its parent and child:
\begin{equation}
    \mathbf{u}_k = \mathbf{p}_{a_k} - \mathbf{p}_{j_k}, \qquad
    \mathbf{v}_k = \mathbf{p}_{c_k} - \mathbf{p}_{j_k},
    \label{eq:displacement}
\end{equation}
The inter-joint angle is then computed via the normalised dot product:
\begin{equation}
    \theta_k = \arccos\!\left(\frac{\mathbf{u}_k \cdot \mathbf{v}_k}{\|\mathbf{u}_k\|\;\|\mathbf{v}_k\|}\right), \quad k \in \{1,\ldots,20\},
    \label{eq:angle}
\end{equation}
where $\|\cdot\|$ denotes the Euclidean norm and $\cdot$ the standard inner product in $\mathbb{R}^3$.
Since there are exactly 20 non-wrist joints in the hand skeleton, this procedure yields the \textbf{angle} representation $\mathbf{x}_{\mathrm{angle}} \in \mathbb{R}^{20}$.
To combine positional and angular cues, we additionally form the \textbf{raw\_angle} concatenation:
\begin{equation}
    \mathbf{x}_{\mathrm{raw\_angle}} = [\mathbf{x}_{\mathrm{raw}};\; \mathbf{x}_{\mathrm{angle}}] \in \mathbb{R}^{83},
    \label{eq:rawangle}
\end{equation}
where $[\,\cdot\,;\,\cdot\,]$ denotes vector concatenation and $83 = 63 + 20$.

\paragraph{Invariance property.}
A central advantage of the angle representation is its provable invariance to similarity transforms.
Let $T\colon \mathbf{p} \mapsto s\,R\,\mathbf{p} + \mathbf{t}$ be a similarity transform with rotation $R \in \mathrm{SO}(3)$ (the special orthogonal group, \ie, the group of $3{\times}3$ rotation matrices), isotropic scale $s > 0$, and translation $\mathbf{t} \in \mathbb{R}^3$.
Translation cancels in the displacement vectors (\cref{eq:displacement}) because it is subtracted out when computing differences, while rotation and scaling cancel in the normalised dot product:
\begin{equation}
    \frac{(s R\,\mathbf{u}_k)\!\cdot\!(s R\,\mathbf{v}_k)}{\|s R\,\mathbf{u}_k\|\;\|s R\,\mathbf{v}_k\|}
    = \frac{s^2\,\mathbf{u}_k^\top R^\top R\,\mathbf{v}_k}{s^2\,\|\mathbf{u}_k\|\;\|\mathbf{v}_k\|}
    = \frac{\mathbf{u}_k \cdot \mathbf{v}_k}{\|\mathbf{u}_k\|\;\|\mathbf{v}_k\|},
    \label{eq:invariance}
\end{equation}
Hence each angle $\theta_k$ is invariant to rotation, translation, and isotropic scaling.
This means the angle representation is \emph{intrinsically} portable across datasets captured under different camera configurations---it requires no normalisation preprocessing whatsoever.
We verify this theoretical prediction empirically in \cref{sec:ablation}.

\paragraph{Relationship to prior angle features.}
Joint-angle descriptors are well established in skeleton-based action recognition~\cite{shi2019twostream,chen2021ctrgcn} and classical hand gesture systems~\cite{cheok2019review}.
Our contribution is not the angle concept per se but rather its specific instantiation and application: (a)~a concrete 20-dimensional formulation defined over the MediaPipe hand topology, (b)~a formal $\mathrm{SO}(3)$-invariance proof accompanied by empirical validation (\cref{sec:ablation}), and (c)~a systematic evaluation demonstrating the effectiveness of these invariant features for cross-lingual few-shot sign-language recognition.

\subsection{Model Architecture}
\label{sec:encoders}

All models follow a two-stage architecture: an encoder $f_\phi$ maps a hand-keypoint vector to a 128-dimensional embedding, and a Prototypical Network head~\cite{snell2017prototypical} classifies queries by nearest-prototype matching in that embedding space.
Here, $\phi$ denotes the trainable parameters of the encoder and $\mathbf{z} = f_\phi(\mathbf{x}) \in \mathbb{R}^{128}$ is the resulting embedding.
We evaluate two encoder backbones; both produce embeddings $\mathbf{z}\!\in\!\mathbb{R}^{128}$.

\paragraph{MLP encoder.}
The input vector (63-dimensional for \texttt{raw}, 20-dimensional for \texttt{angle}, or 83-dimensional for \texttt{raw\_angle}) passes through two fully connected layers of 256 units each.
Each layer applies the sequence \texttt{Linear}$\!\to$\texttt{BatchNorm1d}$\!\to$\texttt{ReLU}$\!\to$\texttt{Dropout}(0.3), followed by a final linear projection to 128~dimensions.
The total parameter count ranges from ${\sim}$105\,k (\texttt{angle}) to ${\sim}$121\,k (\texttt{raw\_angle}); see \cref{tab:architecture}.

\begin{table}[t]
  \caption{Encoder parameter counts by representation. All encoders output 128-D embeddings.}
  \label{tab:architecture}
  \centering\small
  \begin{tabular}{@{}llrr@{}}
    \toprule
    Encoder & Repr & Input Dim & Parameters \\
    \midrule
    MLP & \texttt{raw} & 63 & 116,096 \\
    MLP & \texttt{angle} & 20 & 105,088 \\
    MLP & \texttt{raw\_angle} & 83 & 121,216 \\
    Transformer & \texttt{raw} & 63 & 282,240 \\
    Transformer & \texttt{angle} & 20 & 284,416 \\
    Transformer & \texttt{raw\_angle} & 83 & 292,480 \\
    \bottomrule
  \end{tabular}
\end{table}

\paragraph{Spatial Transformer encoder.}
For the \texttt{raw} representation, the 21 keypoints are treated as a length-21 token sequence with three-dimensional token embeddings, projected to $d_{\mathrm{model}}{=}128$ with sinusoidal positional encodings.
Two Transformer encoder layers (4 heads, FFN dimension 256, dropout 0.1) apply multi-head self-attention; outputs are mean-pooled across the sequence and projected to 128 dimensions with LayerNorm.
For \texttt{angle} and \texttt{raw\_angle} inputs, each angle $\theta_k$ spans a multi-keypoint triplet (\cref{eq:angle}), so no natural per-keypoint tokenisation exists.  Consequently, the full vector is treated as a single token, and self-attention degenerates to a feedforward path, making the Transformer functionally equivalent to a deeper MLP in this case.

\paragraph{Prototypical Network head.}
Given an $N$-way $K$-shot episode, the prototype for each class $n$ is computed as the mean of its $K$ support embeddings: $\mathbf{c}_n = \frac{1}{K}\sum_{i=1}^{K} f_\phi(\mathbf{x}_i^{(n)})$.
A query sample $\mathbf{x}_q$ with embedding $\mathbf{z}_q = f_\phi(\mathbf{x}_q) \in \mathbb{R}^{128}$ is assigned to the class with the nearest prototype: $\hat{y}_q = \arg\min_n \|\mathbf{z}_q - \mathbf{c}_n\|^2$.  Applying a softmax over negative squared distances yields class probabilities.
No learnable parameters are introduced beyond those of the encoder.

\subsection{Few-Shot Evaluation Protocol}
\label{sec:protocol}

\paragraph{Episode definition.}
Each episode samples $N{=}5$ classes from the eligible pool (classes with $\geq K{+}Q$ test samples), draws $K$ support and $Q{=}15$ disjoint query examples per class, and classifies queries by nearest prototype~\cite{snell2017prototypical}.

\paragraph{Within-domain evaluation.}
A randomly initialised encoder is trained and evaluated episodically on each dataset's own test split, without any pretraining on external data.

\paragraph{Cross-lingual transfer evaluation.}
The encoder is first pretrained on the source language's train split using a supervised contrastive loss and then evaluated on each target language under two adaptation modes:
\begin{itemize}[nosep,leftmargin=*]
  \item \emph{Frozen}: all encoder weights are fixed, measuring how well the learned embedding transfers without any adaptation.
  \item \emph{Target-supervised}: the last linear layer is fine-tuned on the target language's train split, quantifying the benefit of minimal adaptation.
\end{itemize}

\paragraph{Reporting.}
All reported results are averaged over 600 episodes with per-episode random seeds and include 95\% confidence intervals.

\subsection{Implementation Details}
\label{sec:implementation}

We release the full implementation as a modular, open-source PyTorch codebase.\footnote{\url{https://github.com/fjkrch/sign_metric_learning}}
The key components are summarised below.

\paragraph{Data pipeline (\texttt{data/}).}
Raw RGB images are processed by MediaPipe Hands v0.10~\cite{zhang2020mediapipe} into per-sample \texttt{.npy} files of shape $(21,3)$, each storing three-dimensional keypoint coordinates for one hand.
A representation layer converts these landmarks on the fly into one of three formats: \texttt{raw} (flattened to 63 dimensions), \texttt{angle} (20 inter-joint angles from anatomical triplets; \cref{eq:angle}), or \texttt{raw\_angle} (concatenation of both, 83 dimensions).
Deterministic stratified 70/30 train/test splits are stored as JSON files in \texttt{splits/} and loaded via \texttt{SplitLandmarkDataset}, which validates zero train--test overlap at initialisation.

\paragraph{Episodic sampling (\texttt{data/episodes.py}).}
A custom \texttt{EpisodicSampler} generates $N$-way $K$-shot episodes, each seeded by \texttt{seed + episode\_index} for exact reproducibility.
Within each episode, $K$ support and $Q$ query samples per class are drawn without replacement, and a \texttt{split\_support\_query} utility partitions the batch while re-labelling classes to $\{0,\ldots,N{-}1\}$.

\paragraph{Encoders (\texttt{models/}).}
The MLP encoder (\texttt{mlp\_encoder.py}) stacks \texttt{Linear}$\!\to\!$\texttt{BatchNorm1d}$\!\to\!$\texttt{ReLU}$\!\to\!$\texttt{Dropout}(0.3) blocks followed by a final 128-D projection.
The Transformer encoder (\texttt{temporal\_transformer.py}) projects each of the 21 keypoints from 3-D to $d_{\mathrm{model}}{=}128$ with sinusoidal positional encodings, applies two self-attention layers (4~heads, FFN~256, dropout~0.1), mean-pools, and projects to 128-D.
Both encoders feed into a shared Prototypical Network head (\texttt{prototypical.py}).
A factory in \texttt{models/\_\_init\_\_.py} routes encoder and model construction by name.

\paragraph{Training and losses (\texttt{train.py}, \texttt{losses/}).}
The training objective combines episode classification loss (negative log-likelihood on ProtoNet log-probabilities) with a supervised contrastive (SupCon) loss~\cite{khosla2020supcon} weighted by 0.5 and using temperature $\tau{=}0.07$.
Optimisation uses AdamW with learning rate $10^{-4}$, weight decay $10^{-4}$, gradient clipping at 1.0, and a cosine annealing schedule.
Early stopping with patience 15 halts training when episode accuracy plateaus.

\paragraph{Cross-lingual adaptation (\texttt{adapt.py}).}
For cross-lingual transfer, a pretrained checkpoint is loaded, the backbone is optionally frozen, and the last linear layer is fine-tuned on the target language's train split for up to 20 epochs (learning rate $10^{-4}$).

\paragraph{Evaluation and reproducibility (\texttt{evaluate.py}, \texttt{tools/}).}
Evaluation scripts compute per-episode accuracy with 95\% confidence intervals ($1.96\,\sigma/\sqrt{n}$) and produce t-SNE visualisations and confusion matrices.
Helper scripts in \texttt{tools/} automate the full experimental matrix (\texttt{run\_full\_matrix.py}), cross-domain evaluation (\texttt{run\_cross\_domain\_expanded.py}), baseline comparisons (\texttt{run\_baselines.py}), and \LaTeX\ table export (\texttt{export\_tables.py}).
All experiments are governed by YAML configuration files in \texttt{configs/}.
Multi-seed robustness checks (seeds 42, 1337, 2024) confirm standard deviations below 1\,pp across all datasets.  
\section{Experiments}
\label{sec:experiments}

We evaluate the proposed geometry-aware representation along four axes: within-domain few-shot performance (\cref{sec:within}), cross-lingual transfer from ASL (\cref{sec:cross}), multi-source transfer (\cref{sec:multisource}), normalisation ablation (\cref{sec:ablation}), and baseline comparison (\cref{sec:baselines}).  We conclude with a targeted analysis of the most challenging dataset, Thai fingerspelling (\cref{sec:thai}).

\subsection{Experimental Setup}
\label{sec:setup}

All experiments run on a consumer CPU (AMD Ryzen 7 7840HS); an NVIDIA GeForce RTX 4060 Laptop GPU is available but not required.
Hand-keypoint extraction uses MediaPipe Hands v0.10~\cite{zhang2020mediapipe}; models are implemented in PyTorch 2.10 with the AdamW optimiser; all runs are fully deterministic (seed~42).
On this hardware, a single training epoch on ASL (${\sim}$44\,k samples) takes approximately 1.5 minutes with the Transformer encoder and under 30 seconds with the MLP; a 600-episode 5-way 5-shot evaluation completes in ${\sim}$7 seconds per dataset.

\subsection{Datasets}
\label{sec:datasets}

\begin{table}[t]
  \caption{Dataset statistics after MediaPipe extraction and 70/30 splitting (seed 42). Eligible = classes with $\geq K{+}Q$ test samples at $K{=}5$, $Q{=}15$.}
  \label{tab:datasets}
  \centering
  \small
  \resizebox{\columnwidth}{!}{%
  \begin{tabular}{@{}lrrrrr@{}}
    \toprule
    Dataset & Classes & Train & Test & Min $n_c^{\text{test}}$ & Elig.\ ($K{=}5$) \\
    \midrule
    ASL Alphabet~\cite{kaggle_asl}     & 29 & 44\,498 & 19\,093 & 1 & 28/29 \\
    LIBRAS Alphabet~\cite{kaggle_libras} & 21 & 23\,993 & 10\,291 & 447 & 21/21 \\
    Arabic SL~\cite{kaggle_arabic}       & 31 &  4\,952 &  2\,141 & 56 & 31/31 \\
    Thai Finger.~\cite{kaggle_thai}      & 42 &  1\,998 &    863 & 12 & 27/42 \\
    \bottomrule
  \end{tabular}}%
\end{table}

We evaluate on four fingerspelling datasets drawn from distinct language families (\cref{tab:datasets}): ASL (American, 29 classes, studio-style RGB)~\cite{kaggle_asl}, LIBRAS (Brazilian, 21 classes, varied backgrounds)~\cite{kaggle_libras}, Arabic SL (31 classes, controlled capture)~\cite{kaggle_arabic}, and Thai Fingerspelling (42 classes, smallest per-class counts)~\cite{kaggle_thai}.
In \cref{tab:datasets}, $\min n_c^{\text{test}}$ denotes the minimum per-class test count, and \emph{Eligible}$(K{=}5)$ indicates the number of classes with $\geq K + Q = 20$ test samples.
All images are processed through MediaPipe Hands; samples for which no hand is detected are discarded.
Splits are deterministic JSON files generated by stratified 70/30 partitioning (seed~42); all reported results use the \textbf{test split} exclusively.

\subsection{Within-Domain Results: How Discriminative Are Angle Features?}
\label{sec:within}

\begin{table*}[t]
  \caption{Within-domain few-shot accuracy (\%) on the \textbf{test split}. 5-way $K$-shot, $Q{=}15$, ProtoNet (Euclidean), 600 episodes, seed 42. Bold = best accuracy per dataset across all representations and encoders.}
  \label{tab:within}
  \centering
  \small
  \begin{tabular}{@{}ll ccc ccc@{}}
    \toprule
    & & \multicolumn{3}{c}{MLP Encoder} & \multicolumn{3}{c}{Transformer Encoder} \\
    \cmidrule(lr){3-5} \cmidrule(lr){6-8}
    Dataset & Repr.& 1-shot & 3-shot & 5-shot & 1-shot & 3-shot & 5-shot \\
    \midrule
    \multirow{3}{*}{ASL}
      & \texttt{raw}        & 88.7\ci{0.7} & 93.8\ci{0.5} & 94.9\ci{0.4}  & 74.2\ci{1.0} & 80.3\ci{0.9} & 82.1\ci{0.9} \\
      & \texttt{angle}      & 81.5\ci{0.9} & 86.8\ci{0.7} & 88.4\ci{0.6}  & 80.8\ci{0.9} & 86.2\ci{0.7} & 87.7\ci{0.7} \\
      & \texttt{raw\_angle} & 90.4\ci{0.7} & 94.5\ci{0.5} & \textbf{95.4}\ci{0.4}  & \textbf{90.8}\ci{0.7} & \textbf{94.7}\ci{0.4} & \textbf{95.4}\ci{0.4} \\
    \midrule
    \multirow{3}{*}{LIBRAS}
      & \texttt{raw}        & 69.7\ci{1.0} & 78.9\ci{0.8} & 81.2\ci{0.8}  & 61.1\ci{0.9} & 65.9\ci{0.9} & 67.9\ci{0.9} \\
      & \texttt{angle}      & \textbf{89.2}\ci{0.7} & \textbf{92.9}\ci{0.5} & \textbf{94.1}\ci{0.5}  & 86.9\ci{0.8} & 91.1\ci{0.6} & 92.5\ci{0.5} \\
      & \texttt{raw\_angle} & 71.6\ci{1.0} & 82.2\ci{0.8} & 84.6\ci{0.8}  & 72.8\ci{1.0} & 82.3\ci{0.8} & 84.7\ci{0.7} \\
    \midrule
    \multirow{3}{*}{Arabic}
      & \texttt{raw}        & 51.3\ci{0.9} & 60.6\ci{0.8} & 64.5\ci{0.8}  & 43.1\ci{0.9} & 47.1\ci{1.0} & 49.0\ci{0.9} \\
      & \texttt{angle}      & \textbf{81.1}\ci{0.9} & \textbf{88.2}\ci{0.6} & \textbf{89.8}\ci{0.6}  & 78.3\ci{0.9} & 85.7\ci{0.7} & 87.8\ci{0.6} \\
      & \texttt{raw\_angle} & 55.5\ci{1.0} & 66.6\ci{0.9} & 71.0\ci{0.9}  & 55.6\ci{1.0} & 63.2\ci{0.9} & 67.3\ci{0.9} \\
    \midrule
    \multirow{3}{*}{Thai}
      & \texttt{raw}        & 40.3\ci{0.7} & 47.0\ci{0.7} & 48.8\ci{0.8}  & 33.3\ci{0.7} & 35.7\ci{0.6} & 36.0\ci{0.6} \\
      & \texttt{angle}      & \textbf{46.2}\ci{0.8} & \textbf{51.2}\ci{0.8} & \textbf{52.7}\ci{0.8}  & 44.6\ci{0.8} & 50.6\ci{0.8} & 51.8\ci{0.8} \\
      & \texttt{raw\_angle} & 43.4\ci{0.8} & 50.2\ci{0.8} & 51.9\ci{0.8}  & 42.3\ci{0.7} & 47.1\ci{0.7} & 48.7\ci{0.8} \\
    \bottomrule
  \end{tabular}
\end{table*}

\Cref{tab:within} reports within-domain few-shot accuracy, where each model is both trained and evaluated on the same dataset.
Rows are grouped by dataset; columns compare three input representations across two encoder architectures.

\textbf{Key findings.}
On LIBRAS, Arabic, and Thai, the \texttt{angle} representation paired with the MLP encoder consistently achieves the highest accuracy.  The largest gains over normalised \texttt{raw} coordinates appear on Arabic (+25.3 percentage points at 5-shot) and LIBRAS (+12.9 percentage points).
On ASL---the largest dataset---the \texttt{raw\_angle} concatenation slightly outperforms individual representations (95.4\%).  This is likely because the abundant training data allows the network to exploit complementary coordinate and angle cues, whereas on smaller datasets the additional dimensionality introduces noise that outweighs the benefit.
The MLP encoder generally matches or outperforms the Transformer, a finding consistent with Chen~\etal~\cite{chen2019closer}: simple encoders suffice when the input representation is well designed.

\subsection{Cross-Lingual Transfer from ASL: Does Geometric Invariance Reduce Domain Shift?}
\label{sec:cross}

\begin{table*}[t]
  \caption{Effect of ASL pretraining on few-shot accuracy (\%): encoder pretrained on ASL (SupCon), evaluated on each target's \textbf{test split}. 5-way 5-shot, $Q{=}15$, 600 episodes, seed 42. Bold = best accuracy within each (target, mode) pair across all representations and encoders. ASL$\to$ASL rows are in-domain references (source{=}target), not cross-lingual transfer.\textsuperscript{$\dagger$}}
  \label{tab:cross}
  \centering
  \small
  \begin{tabular}{@{}ll ccc ccc@{}}
    \toprule
    & & \multicolumn{3}{c}{MLP Encoder} & \multicolumn{3}{c}{Transformer Encoder} \\
    \cmidrule(lr){3-5} \cmidrule(lr){6-8}
    Target & Mode & \texttt{raw} & \texttt{angle} & \texttt{raw\_angle} & \texttt{raw} & \texttt{angle} & \texttt{raw\_angle} \\
    \midrule
    \multirow{2}{*}{ASL}
      & Frozen       & 97.0\ci{0.4} & 91.7\ci{0.5} & \textbf{97.1}\ci{0.4} & 92.7\ci{0.6} & 87.5\ci{0.7} & \textbf{97.1}\ci{0.3} \\
      & Target-sup.  & 97.0\ci{0.4} & 91.7\ci{0.5} & \textbf{97.1}\ci{0.4} & 92.7\ci{0.6} & 87.5\ci{0.7} & \textbf{97.1}\ci{0.3} \\
    \midrule
    \multirow{2}{*}{LIBRAS}
      & Frozen       & 86.5\ci{0.8} & \textbf{95.0}\ci{0.4} & 87.8\ci{0.7} & 81.3\ci{0.8} & 93.2\ci{0.5} & 88.3\ci{0.7} \\
      & Target-sup.  & 94.2\ci{0.5} & \textbf{96.0}\ci{0.4} & 95.2\ci{0.5} & 81.4\ci{0.8} & 93.2\ci{0.5} & 88.4\ci{0.7} \\
    \midrule
    \multirow{2}{*}{Arabic}
      & Frozen       & 74.2\ci{0.8} & \textbf{91.3}\ci{0.5} & 76.6\ci{0.9} & 66.1\ci{0.9} & 87.4\ci{0.7} & 75.9\ci{0.9} \\
      & Target-sup.  & 89.4\ci{0.6} & 92.7\ci{0.5} & \textbf{92.9}\ci{0.5} & 66.2\ci{0.9} & 87.4\ci{0.7} & 76.2\ci{0.8} \\
    \midrule
    \multirow{2}{*}{Thai}
      & Frozen       & 52.5\ci{0.9} & 53.2\ci{0.8} & 50.6\ci{0.8} & 48.7\ci{0.8} & 48.8\ci{0.8} & \textbf{53.8}\ci{0.8} \\
      & Target-sup.  & \textbf{58.5}\ci{0.9} & 57.4\ci{0.8} & 57.3\ci{0.8} & 48.8\ci{0.8} & 48.8\ci{0.8} & 53.9\ci{0.8} \\
    \bottomrule
    \multicolumn{8}{@{}l@{}}{\rule{0pt}{2.2ex}\textsuperscript{$\dagger$}\footnotesize ASL$\to$ASL (source{=}target) is an in-domain reference, not cross-lingual transfer.}
  \end{tabular}
\end{table*}

\Cref{tab:cross} reports few-shot accuracy when the encoder is pretrained on ASL; the ASL$\to$ASL rows serve as in-domain references (source equals target).

\textbf{Key findings.}
With a frozen MLP encoder, \texttt{angle} features achieve 95.0\% on LIBRAS and 91.3\% on Arabic---surpassing \texttt{raw} coordinates by 8.5 and 17.1 percentage points, respectively---confirming that geometric invariance substantially reduces cross-dataset domain shift.
Target-supervised fine-tuning further improves these results to 96.0\% and 92.7\%.
Notably, the best ASL$\to$Thai result (58.5\%) exceeds the within-domain baseline (52.7\%; \cref{tab:within}), demonstrating that source-language pretraining benefits even the most challenging target.
The consistent target difficulty ordering---LIBRAS~$>$~Arabic~$>$~Thai---holds across all representations and adaptation modes.

\subsection{Multi-Source Transfer: Which Source Language Works Best?}
\label{sec:multisource}

\begin{table}[t]
  \caption{Multi-source transfer: LIBRAS$\leftrightarrow$Arabic per-representation breakdown (MLP, 5-way 5-shot, $Q{=}15$, 600 episodes, seed 42). Bold = best representation within each (direction, mode) pair.}
  \label{tab:multisource}
  \centering\small
  \begin{tabular}{@{}ll cc@{}}
    \toprule
    Direction & Repr & Frozen & Target-sup. \\
    \midrule
    \multirow{3}{*}{LIBRAS$\to$Arabic}
      & \texttt{raw}        & 89.8\ci{0.6} & 91.8\ci{0.6} \\
      & \texttt{angle}      & 91.2\ci{0.6} & 92.2\ci{0.5} \\
      & \texttt{raw\_angle} & \textbf{91.7}\ci{0.6} & \textbf{93.8}\ci{0.5} \\
    \midrule
    \multirow{3}{*}{Arabic$\to$LIBRAS}
      & \texttt{raw}        & 95.9\ci{0.5} & 96.5\ci{0.4} \\
      & \texttt{angle}      & 94.9\ci{0.5} & 95.9\ci{0.4} \\
      & \texttt{raw\_angle} & \textbf{97.1}\ci{0.4} & \textbf{97.4}\ci{0.4} \\
    \bottomrule
  \end{tabular}
\end{table}

\begin{table}[t]
  \caption{Best accuracy (\%) by source language (MLP, 5-way 5-shot, $Q{=}15$, 600 episodes, seed 42).  Off-diagonal: frozen-encoder transfer; each cell reports the highest-scoring representation (parenthesised: \texttt{r}=\texttt{raw}, \texttt{a}=\texttt{angle}, \texttt{ra}=\texttt{raw\_angle}).  Diagonal\textsuperscript{$\dagger$}: within-domain baseline (no pretraining; from \cref{tab:within}).  Bold = best cross-lingual source for each target row.}
  \label{tab:source_compare}
  \centering\small
  \resizebox{\columnwidth}{!}{%
  \begin{tabular}{@{}l cccc@{}}
    \toprule
    Target & ASL & LIBRAS & Arabic & Thai \\
    \midrule
    ASL    & 95.4\ci{0.4}\textsuperscript{$\dagger$}\,(\texttt{ra}) & 93.7\ci{0.5}\,(\texttt{r}) & 94.1\ci{0.4}\,(\texttt{r}) & \textbf{95.1}\ci{0.4}\,(\texttt{ra}) \\
    LIBRAS & 95.0\ci{0.4}\,(\texttt{a}) & 94.1\ci{0.5}\textsuperscript{$\dagger$}\,(\texttt{a}) & \textbf{97.1}\ci{0.4}\,(\texttt{ra}) & 94.6\ci{0.5}\,(\texttt{a}) \\
    Arabic & 91.3\ci{0.5}\,(\texttt{a}) & \textbf{91.7}\ci{0.6}\,(\texttt{ra}) & 89.8\ci{0.6}\textsuperscript{$\dagger$}\,(\texttt{a}) & 90.2\ci{0.5}\,(\texttt{a}) \\
    Thai   & 53.2\ci{0.8}\,(\texttt{a}) & 52.2\ci{0.8}\,(\texttt{ra}) & \textbf{54.6}\ci{0.9}\,(\texttt{r}) & 52.7\ci{0.8}\textsuperscript{$\dagger$}\,(\texttt{a}) \\
    \bottomrule
    \multicolumn{5}{@{}l@{}}{\rule{0pt}{2.2ex}\textsuperscript{$\dagger$}\footnotesize Within-domain (no pretraining); all other cells use frozen cross-lingual transfer.}
  \end{tabular}}%
\end{table}

\Cref{tab:source_compare} provides a comprehensive view of all source--target transfer pairs: off-diagonal cells report frozen cross-lingual transfer, while diagonal cells show the within-domain baseline (no pretraining; from \cref{tab:within}), enabling direct comparison.

\textbf{Key findings.}
Source pretraining frequently outperforms target-only few-shot learning: Arabic$\to$LIBRAS reaches 97.1\%, exceeding the within-domain score (94.1\%) by 3.0~percentage points; LIBRAS$\to$Arabic (91.7\%) exceeds within-domain (89.8\%) by 1.9~percentage points; and Arabic$\to$Thai (54.6\%) exceeds within-domain (52.7\%) by 1.9~percentage points.
\Cref{tab:multisource} provides the LIBRAS$\leftrightarrow$Arabic per-representation breakdown.
The optimal source language is target-dependent: Arabic is best for LIBRAS (97.1\%); LIBRAS for Arabic (91.7\%); Arabic for Thai (54.6\%); and Thai for ASL (95.1\%).
The optimal representation also varies with the transfer direction: \texttt{angle} dominates under large domain shift (e.g., ASL$\to$Arabic), \texttt{raw} when the target features standardised capture conditions (ASL), and \texttt{raw\_angle} for LIBRAS$\leftrightarrow$Arabic (\cref{tab:multisource}).
The LIBRAS$\leftrightarrow$Arabic asymmetry may partly reflect differences in dataset scale: LIBRAS provides a richer pretraining signal (23\,993 vs.\ 4\,952 train samples; \cref{tab:datasets}) with more varied backgrounds.
Across both non-ASL transfer directions, \texttt{raw\_angle} is the best or tied-best representation (\cref{tab:multisource}), suggesting that combining positional and angular cues compensates for moderate domain shift when neither component alone suffices.

\subsection{Normalisation Ablation: Are Angles Truly Invariant?}
\label{sec:ablation}

\begin{table}[t]
  \caption{Effect of removing wrist-centring and scale normalisation.  5-way 5-shot, $Q{=}15$, MLP encoder, 600 episodes, seed 42.  ``No norm'' = MediaPipe raw output.}
  \label{tab:ablation}
  \centering
  \small
  \begin{tabular}{@{}llccc@{}}
    \toprule
    Dataset & Repr & No Norm & Normalised & Diff.\ (p.p.) \\
    \midrule
    LIBRAS & \texttt{raw}   & 76.4 & 81.2 & 4.8 \\
    LIBRAS & \texttt{angle} & \textbf{94.4} & \textbf{94.1} & $-$0.3$^\dagger$ \\
    Arabic & \texttt{raw}   & 59.1 & 64.5 & 5.4 \\
    Arabic & \texttt{angle} & \textbf{90.1} & \textbf{89.8} & $-$0.3$^\dagger$ \\
    \bottomrule
  \end{tabular}
  \vspace{2pt}
  \par\noindent{\footnotesize $^\dagger$Within noise band; confirms theoretical invariance (\cref{eq:invariance}).}
\end{table}

\Cref{tab:ablation} provides empirical verification of the theoretical invariance claim: removing both wrist-centring and scale normalisation degrades \texttt{raw} coordinates by 4.8--5.4 percentage points, whereas \texttt{angle} features remain virtually unchanged ($|\Delta| \leq 0.3$ percentage points), confirming the prediction of \cref{eq:invariance}.
A more granular decomposition on Arabic (\cref{tab:pipeline}) reveals the contribution of each processing stage:

\begin{table}[t]
  \caption{Cumulative normalisation ablation on Arabic (MLP / raw).  5-way $K$-shot, $Q{=}15$, 600 episodes, seed 42.}
  \label{tab:pipeline}
  \centering
  \small
  \begin{tabular}{@{}lccc@{}}
    \toprule
    Setting & 1-shot & 3-shot & 5-shot \\
    \midrule
    No normalisation          & 41.8 & 56.3 & 59.1 \\
    + Wrist-centring \& scale & 51.3 & 60.6 & 64.5 \\
    + Geometry-aware (angle)  & \textbf{81.1} & \textbf{88.2} & \textbf{89.8} \\
    \bottomrule
  \end{tabular}
\end{table}

The geometry-aware angle representation provides the single largest improvement---a mean absolute gain of 25.3 percentage points at 5-shot over normalised \texttt{raw} coordinates---demonstrating that the benefit extends well beyond preprocessing to the construction of a more discriminative feature space.

\subsection{Baselines and Robustness: How Much Does the Encoder Help?}
\label{sec:baselines}

To disentangle the encoder's contribution from that of the representation itself, \cref{tab:baselines_ext} compares three 5-way 5-shot classification methods: (i)~\emph{input-space} nearest prototype (no encoder; Euclidean distance computed directly in the raw feature space), (ii)~\emph{episode-linear} (per-episode logistic regression fitted on MLP support embeddings), and (iii)~our ProtoNet with an MLP encoder.

\begin{table}[t]
  \caption{Baseline comparison (5-way 5-shot, $Q{=}15$, 600 episodes, seed 42, MLP encoder where applicable).}
  \label{tab:baselines_ext}
  \centering\small
  \resizebox{\columnwidth}{!}{\begin{tabular}{@{}llccc@{}}
\toprule
Dataset & Repr & Input-Sp. & Ep.-Linear & ProtoNet \\
\midrule
ASL & \texttt{raw} & 95.1\ci{0.4} & 91.9\ci{0.5} & 94.9\ci{0.4} \\
ASL & \texttt{angle} & 89.4\ci{0.6} & 85.7\ci{0.7} & 88.4\ci{0.6} \\
LIBRAS & \texttt{raw} & 81.0\ci{0.8} & 80.2\ci{0.8} & 81.2\ci{0.8} \\
LIBRAS & \texttt{angle} & 94.2\ci{0.5} & 93.1\ci{0.5} & 94.1\ci{0.5} \\
Arabic & \texttt{raw} & 65.0\ci{0.9} & 63.7\ci{0.9} & 64.5\ci{0.8} \\
Arabic & \texttt{angle} & 90.5\ci{0.6} & 88.2\ci{0.6} & 89.8\ci{0.6} \\
Thai & \texttt{raw} & 47.6\ci{0.8} & 50.9\ci{0.8} & 48.8\ci{0.8} \\
Thai & \texttt{angle} & 52.8\ci{0.8} & 51.9\ci{0.8} & 52.7\ci{0.8} \\
\bottomrule
\end{tabular}}
\end{table}

\textbf{Key findings.}
The input-space baseline is competitive with ProtoNet on several datasets (e.g., ASL/raw: 95.1\% vs.\ 94.9\%), confirming that hand-keypoint features are inherently discriminative.
A full-data linear classifier attains 99.7\% on LIBRAS and 93.8\% on ASL; the gap to 5-shot ProtoNet (5.6\,pp on LIBRAS) quantifies the cost of learning from only five examples.
On Thai, ProtoNet slightly exceeds the full-data classifier (52.7\% vs.\ 51.7\%), likely because the limited number of training samples per class (${\sim}$48) constrains the linear model.
Multi-seed evaluation (seeds 42, 1337, 2024) yields standard deviations below 1\,pp on all datasets, confirming the robustness of these results.

\subsection{Analysis: Why Is Thai Fingerspelling Difficult?}
\label{sec:thai}

Thai yields the lowest accuracy among all four datasets.  Three factors contribute to this difficulty: the high class cardinality (42 classes), the small dataset size (${\sim}$2\,860 samples after filtering), and the existence of consonant groups with similar hand shapes.
Despite these challenges, \texttt{angle} features outperform \texttt{raw} by 3.9\,pp at 5-shot (\cref{tab:within}), and ASL$\to$Thai transfer (53.2\%; \cref{tab:source_compare}) slightly exceeds the within-domain baseline (52.7\%).

\begin{figure}[t]
  \centering
  \includegraphics[width=\columnwidth]{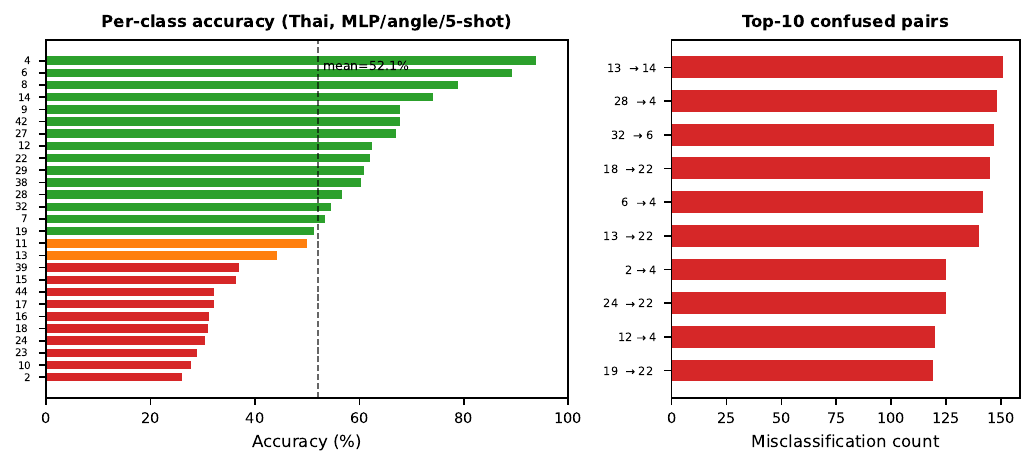}
  \caption{Error analysis on Thai (MLP / angle / 5-way 5-shot, 600 episodes).  \textbf{Left:} Per-class accuracy (26--94\%).  \textbf{Right:} Top confused pairs.}
  \label{fig:thai_errors}
\end{figure}

\Cref{fig:thai_errors} shows per-class accuracy spanning 26--94\%; confusions concentrate on morphologically similar consonant pairs (e.g., classes 13$\to$14, 28$\to$4), suggesting that finer-grained features---such as bone-length ratios or contact information---may be needed to disambiguate these cases.

\section{Discussion}
\label{sec:discussion}

\paragraph{Why invariant features help few-shot transfer.}
In Prototypical Networks, each class centroid is a sample mean over only $K$ support examples.  When the input representation carries extrinsic variance---due to camera viewpoint, hand scale, or global position---the centroid estimate is corrupted by factors unrelated to hand shape, and classification degrades.  The $\mathrm{SO}(3)$-invariant angle representation (\cref{eq:invariance}) eliminates these nuisance factors at the representation level, compressing hand-pose information into a 20-dimensional space where class prototypes are geometrically tighter.  This is particularly critical in the few-shot regime ($K{=}5$), where there is insufficient data to learn invariance from examples alone (\cref{tab:pipeline}).
The empirical consequence is that pretraining on any source language can outperform target-only few-shot training---for example, Arabic$\to$LIBRAS with a frozen encoder exceeds within-domain accuracy by 3.0\,pp (\cref{tab:source_compare})---indicating that the invariant embedding captures portable geometric structure that transfers across sign languages.

\paragraph{When do coordinate features still help?}
On ASL, the largest and most uniformly captured dataset, the \texttt{raw\_angle} concatenation achieves the best results (\cref{tab:within,tab:cross}).  This suggests that when domain shift is small and data are abundant, absolute positional cues provide complementary discriminative information that angles alone discard.  For LIBRAS$\leftrightarrow$Arabic transfer, \texttt{raw\_angle} likewise outperforms \texttt{angle} alone (\cref{tab:multisource}), indicating that moderate domain shift does not fully negate the value of coordinate features.

\paragraph{Limitations.}
Several limitations should be noted.
First, our evaluation covers \emph{static fingerspelling only}; dynamic signs and continuous signing involve temporal trajectories that require sequential modelling.
Second, we use \emph{single-hand keypoints}: two-handed signs and signs involving body or facial cues are outside the current scope.
Third, the angle representation \emph{discards absolute bone-length information}, which could be discriminative for signs that differ primarily in finger extension magnitude or hand size.
Finally, although we evaluate four typologically diverse languages, broader coverage---especially of non-fingerspelling sign lexicons---is needed to assess generality.

\section{Conclusion}
\label{sec:conclusion}

Scaling sign-language recognition to the majority of the world's 300{+} sign languages demands methods for cross-lingual few-shot transfer from data-rich source languages.
We introduced a 20-dimensional $\mathrm{SO}(3)$-invariant joint-angle descriptor that removes a dominant source of cross-dataset domain shift and stabilises Prototypical Network classification across languages.
Evaluated under a unified, deterministic episodic protocol across four typologically diverse fingerspelling alphabets, angle-based features consistently improve over normalised-coordinate baselines, with the largest gains observed under substantial domain shift.
The \texttt{raw\_angle} concatenation frequently emerges as the strongest cross-lingual representation, although the optimal choice remains direction-dependent.
These results demonstrate that formally invariant hand-geometry descriptors can serve as a portable and effective foundation for few-shot sign-language recognition in low-resource settings.  
{
    \small
    \bibliographystyle{ieeenat_fullname}
    \bibliography{main}

\begin{thebibliography}{29}
\providecommand{\natexlab}[1]{#1}
\providecommand{\url}[1]{\texttt{#1}}
\expandafter\ifx\csname urlstyle\endcsname\relax
  \providecommand{\doi}[1]{doi: #1}\else
  \providecommand{\doi}{doi: \begingroup \urlstyle{rm}\Url}\fi

\bibitem[Al-Brham(2023)]{kaggle_arabic}
Muhammad Al-Brham.
\newblock {RGB} {Arabic} alphabets sign language dataset, 2023.
\newblock Kaggle,
  \url{https://www.kaggle.com/datasets/muhammadalbrham/rgb-arabic-alphabets-sign-language-dataset}.

\bibitem[Alani and Cosma(2021)]{alani2021arabic}
Ahmed~A. Alani and Georgina Cosma.
\newblock {ArSL-CNN}: A convolutional neural network for arabic sign language
  gesture recognition.
\newblock In \emph{Proceedings of the International Joint Conference on Neural
  Networks (IJCNN)}, pages 1--8, 2021.

\bibitem[Bilge et~al.(2024)Bilge, Ikizler-Cinbis, and Cinbis]{bilge2024cross}
Yunus~Can Bilge, Nazli Ikizler-Cinbis, and Ramazan~Gokberk Cinbis.
\newblock Towards zero-shot sign language recognition.
\newblock \emph{Pattern Recognition}, 150:\penalty0 110351, 2024.

\bibitem[Boh\'{a}\v{c}ek(2023)]{bohacek2023pose}
Maty\'{a}\v{s} Boh\'{a}\v{c}ek.
\newblock Pose-based few-shot sign language recognition.
\newblock \emph{arXiv preprint arXiv:2301.03769}, 2023.

\bibitem[Camg\"{o}z et~al.(2020)Camg\"{o}z, Koller, Hadfield, and
  Bowden]{camgoz2020sign}
Necati~Cihan Camg\"{o}z, Oscar Koller, Simon Hadfield, and Richard Bowden.
\newblock Sign language transformers: Joint end-to-end sign language
  recognition and translation.
\newblock In \emph{Proceedings of the IEEE/CVF Conference on Computer Vision
  and Pattern Recognition (CVPR)}, pages 10023--10033, 2020.

\bibitem[Chen et~al.(2019)Chen, Liu, Kira, Wang, and Huang]{chen2019closer}
Wei-Yu Chen, Yen-Cheng Liu, Zsolt Kira, Yu-Chiang~Frank Wang, and Jia-Bin
  Huang.
\newblock A closer look at few-shot classification.
\newblock In \emph{Proceedings of the International Conference on Learning
  Representations (ICLR)}, 2019.

\bibitem[Chen et~al.(2021)Chen, Zhang, Yuan, Li, Deng, and Hu]{chen2021ctrgcn}
Yuxin Chen, Ziqi Zhang, Chunfeng Yuan, Bing Li, Ying Deng, and Weiming Hu.
\newblock Channel-wise topology refinement graph convolution for skeleton-based
  action recognition.
\newblock In \emph{Proceedings of the IEEE/CVF International Conference on
  Computer Vision (ICCV)}, pages 13026--13035, 2021.

\bibitem[Cheok et~al.(2019)Cheok, Omar, and Jaward]{cheok2019review}
Ming~Jin Cheok, Zaid Omar, and Mohamed~Hasan Jaward.
\newblock A review of hand gesture and sign language recognition techniques.
\newblock \emph{International Journal of Machine Learning and Cybernetics},
  10\penalty0 (1):\penalty0 131--153, 2019.

\bibitem[Conneau et~al.(2020)Conneau, Khandelwal, Goyal, Chaudhary, Wenzek,
  Guzm{\'a}n, Grave, Ott, Zettlemoyer, and Stoyanov]{conneau2020xlmr}
Alexis Conneau, Kartikay Khandelwal, Naman Goyal, Vishrav Chaudhary, Guillaume
  Wenzek, Francisco Guzm{\'a}n, Edouard Grave, Myle Ott, Luke Zettlemoyer, and
  Veselin Stoyanov.
\newblock Unsupervised cross-lingual representation learning at scale.
\newblock In \emph{Proceedings of the Annual Meeting of the Association for
  Computational Linguistics (ACL)}, pages 8440--8451, 2020.

\bibitem[de~Amorim et~al.(2023)de~Amorim, Mac{\^e}do, and
  Zanchettin]{de2023sign}
Cl{\'a}udio~Carvalho de Amorim, David Mac{\^e}do, and Cleber Zanchettin.
\newblock Sign language recognition using skeleton data with {MediaPipe}.
\newblock In \emph{Proceedings of the International Joint Conference on Neural
  Networks (IJCNN)}, pages 1--8, 2023.

\bibitem[de~Oliveira(2022)]{kaggle_libras}
Willian~Soares de Oliveira.
\newblock {LIBRAS}: {Brazilian Sign Language} alphabet dataset, 2022.
\newblock Kaggle,
  \url{https://www.kaggle.com/datasets/williansoliveira/libras}.

\bibitem[{Ethnologue}(2024)]{ethnologue2024}
{Ethnologue}.
\newblock Sign languages of the world, 2024.
\newblock \url{https://www.ethnologue.com}.

\bibitem[Finn et~al.(2017)Finn, Abbeel, and Levine]{finn2017maml}
Chelsea Finn, Pieter Abbeel, and Sergey Levine.
\newblock Model-agnostic meta-learning for fast adaptation of deep networks.
\newblock In \emph{Proceedings of the International Conference on Machine
  Learning (ICML)}, pages 1126--1135, 2017.

\bibitem[Hartmann(2023)]{kaggle_thai}
Nicki Hartmann.
\newblock Thai letter sign language dataset, 2023.
\newblock Kaggle,
  \url{https://www.kaggle.com/datasets/nickihartmann/thai-letter-sign-language}.

\bibitem[Jiang et~al.(2021)Jiang, Sun, Wang, Bai, Li, and
  Fu]{jiang2021skeleton}
Songyao Jiang, Bin Sun, Ling Wang, Yue Bai, Kunpeng Li, and Yun Fu.
\newblock Skeleton aware multi-modal sign language recognition.
\newblock In \emph{Proceedings of the IEEE/CVF Conference on Computer Vision
  and Pattern Recognition Workshops (CVPRW)}, pages 3413--3423, 2021.

\bibitem[Khosla et~al.(2020)Khosla, Teterwak, Wang, Sarna, Tian, Isola,
  Maschinot, Liu, and Krishnan]{khosla2020supcon}
Prannay Khosla, Piotr Teterwak, Chen Wang, Aaron Sarna, Yonglong Tian, Phillip
  Isola, Aaron Maschinot, Ce Liu, and Dilip Krishnan.
\newblock Supervised contrastive learning.
\newblock In \emph{Advances in Neural Information Processing Systems
  (NeurIPS)}, pages 18661--18673, 2020.

\bibitem[Li et~al.(2020)Li, Rodriguez-Opazo, Yu, and Li]{li2020word}
Dongxu Li, Cristian Rodriguez-Opazo, Xin Yu, and Hongdong Li.
\newblock Word-level deep sign language recognition from video: A new
  large-scale dataset and methods comparison.
\newblock In \emph{Proceedings of the IEEE/CVF Winter Conference on
  Applications of Computer Vision (WACV)}, pages 1459--1469, 2020.

\bibitem[Li et~al.(2022)Li, Wang, Tang, Tran, Tang, Juan, Baevski, and
  Auli]{li2022massively}
Xian Li, Changhan Wang, Yun Tang, Chau Tran, Yuqing Tang, Dieysa Juan, Alexei
  Baevski, and Michael Auli.
\newblock Massively multilingual {ASR}: 50 languages, 1 model, 1 billion
  parameters.
\newblock In \emph{Proceedings of Interspeech}, pages 4751--4755, 2022.

\bibitem[Mavi(2020)]{mavi2020asl}
Akash Mavi.
\newblock A real-time applicable 3d hand gesture recognition system using deep
  learning with {ASL} dataset.
\newblock \emph{International Journal of Intelligent Systems and Applications
  in Engineering}, 8\penalty0 (4):\penalty0 204--212, 2020.

\bibitem[Nagaraj(2018)]{kaggle_asl}
Akash Nagaraj.
\newblock {ASL} alphabet: Image dataset for alphabets in {American Sign
  Language}, 2018.
\newblock Kaggle,
  \url{https://www.kaggle.com/datasets/grassknoted/asl-alphabet}.

\bibitem[Podder et~al.(2022)Podder, Chowdhury, Tahir, Mahbub, Khandakar,
  Hossain, and Abbas]{podder2022libras}
Kaushik~Kumar Podder, Muhammad E.~H. Chowdhury, Anas~M. Tahir, Zaid~Bin Mahbub,
  Amith Khandakar, Md.~Shafayet Hossain, and Tawil~O. Abbas.
\newblock Bangla sign language ({BdSL}) alphabets \& numerals classification
  using a deep learning model.
\newblock \emph{Sensors}, 22\penalty0 (2):\penalty0 574, 2022.

\bibitem[Shi et~al.(2019)Shi, Zhang, Cheng, and Lu]{shi2019twostream}
Lei Shi, Yifan Zhang, Jian Cheng, and Hanqing Lu.
\newblock Two-stream adaptive graph convolutional networks for skeleton-based
  action recognition.
\newblock In \emph{Proceedings of the IEEE/CVF Conference on Computer Vision
  and Pattern Recognition (CVPR)}, pages 12026--12035, 2019.

\bibitem[Shin et~al.(2023)Shin, Musa, and Hasan]{shin2023korean}
Jungpil Shin, Md. Al~Mehedi Musa, and Md. Al~Mehedi Hasan.
\newblock Korean sign language recognition using {MediaPipe} and deep learning.
\newblock In \emph{International Conference on Big Data and Smart Computing
  (BigComp)}, pages 365--368, 2023.

\bibitem[Snell et~al.(2017)Snell, Swersky, and Zemel]{snell2017prototypical}
Jake Snell, Kevin Swersky, and Richard~S. Zemel.
\newblock Prototypical networks for few-shot learning.
\newblock In \emph{Advances in Neural Information Processing Systems
  (NeurIPS)}, pages 4077--4087, 2017.

\bibitem[Sung et~al.(2018)Sung, Yang, Zhang, Xiang, Torr, and
  Hospedales]{sung2018relation}
Flood Sung, Yongxin Yang, Li Zhang, Tao Xiang, Philip H.~S. Torr, and
  Timothy~M. Hospedales.
\newblock Learning to compare: Relation network for few-shot learning.
\newblock In \emph{Proceedings of the IEEE/CVF Conference on Computer Vision
  and Pattern Recognition (CVPR)}, pages 1199--1208, 2018.

\bibitem[Tavella et~al.(2022)Tavella, Schlegel, Romeo, Kola{\v{c}}ek, and
  Caramiaux]{tavella2022fewshot}
Federico Tavella, Viktor Schlegel, Marta Romeo, Martin Kola{\v{c}}ek, and
  Baptiste Caramiaux.
\newblock Few-shot sign language recognition with transfer learning.
\newblock In \emph{Proceedings of the LREC Workshop on Resources and
  Technologies for Indigenous, Endangered and Lesser-resourced Languages},
  pages 94--100, 2022.

\bibitem[Vinyals et~al.(2016)Vinyals, Blundell, Lillicrap, Kavukcuoglu, and
  Wierstra]{vinyals2016matching}
Oriol Vinyals, Charles Blundell, Timothy Lillicrap, Koray Kavukcuoglu, and Daan
  Wierstra.
\newblock Matching networks for one shot learning.
\newblock In \emph{Advances in Neural Information Processing Systems
  (NeurIPS)}, pages 3630--3638, 2016.

\bibitem[{World Health Organization}(2024)]{who2024deafness}
{World Health Organization}.
\newblock Deafness and hearing loss, 2024.
\newblock Fact sheet,
  \url{https://www.who.int/news-room/fact-sheets/detail/deafness-and-hearing-loss}.

\bibitem[Zhang et~al.(2020)Zhang, Bazarevsky, Vakunov, Tkachenka, Sung, Chang,
  and Grundmann]{zhang2020mediapipe}
Fan Zhang, Valentin Bazarevsky, Andrey Vakunov, Andrei Tkachenka, George Sung,
  Chuo-Ling Chang, and Matthias Grundmann.
\newblock {MediaPipe Hands}: On-device real-time hand tracking.
\newblock \emph{arXiv preprint arXiv:2006.10214}, 2020.

\end{thebibliography}
}

\end{document}